\documentclass[11pt]{article}

\usepackage[final]{acl}

\usepackage{times}
\usepackage{latexsym}

\usepackage[T1]{fontenc}

\usepackage[utf8]{inputenc}

\usepackage{microtype}

\usepackage{inconsolata}

\usepackage{graphicx}
\usepackage{booktabs}
\usepackage{multirow}
\usepackage{enumitem}
\usepackage{amsmath}
\usepackage{algorithm}
\usepackage{algpseudocode}
\usepackage{hyperref}

\title{EvoTaxo: Building and Evolving Taxonomy from \\ Social Media Streams}

\author{
 \textbf{Yiyang Li\textsuperscript{1}},
 \textbf{Tianyi Ma\textsuperscript{1}},
 \textbf{Yanfang Ye\textsuperscript{1\textdagger}}
\\
 \textsuperscript{1}University of Notre Dame \quad \textsuperscript{\textdagger}Corresponding Author \\
 \texttt{\{yli62, yye7\}@nd.edu}
}

\begin{document}
\maketitle
\begin{abstract}
Constructing taxonomies from social media corpora is challenging because posts are short, noisy, semantically entangled, and temporally dynamic. Existing taxonomy induction methods are largely designed for static corpora and often struggle to balance robustness, scalability, and sensitivity to evolving discourse. We propose \textbf{EvoTaxo}, a LLM-based framework for building and evolving taxonomies from temporally ordered social media streams. Rather than clustering raw posts directly, EvoTaxo converts each post into a structured draft action over the current taxonomy, accumulates structural evidence over time windows, and consolidates candidate edits through dual-view clustering that combines semantic similarity with temporal locality. A refinement-and-arbitration procedure then selects reliable edits before execution, while each node maintains a concept memory bank to preserve semantic boundaries over time. Experiments on two Reddit corpora show that EvoTaxo produces more balanced taxonomies than baselines, with clearer post-to-leaf assignment, better corpus coverage at comparable taxonomy size, and stronger structural quality. A case study on the Reddit community \texttt{/r/ICE\_Raids} further shows that EvoTaxo captures meaningful temporal shifts in discourse. Our codebase is available \href{https://github.com/Yiyang-Ian-Li/EvoTaxo}{here}.
\end{abstract}

\section{Introduction}

Social media platforms have become a major arena for public discourse, providing continuous records of opinions, reactions, and collective narratives at an unprecedented scale~\cite{dong2021review, deng2021cross, li2024pro, zhu2025ratsd}. These data have supported a wide range of computational studies on public opinion, political communication, and social dynamics~\cite{lazer2009computational, zhuravskaya2020political, reveilhac2022systematic}. Yet the resulting discourse space is difficult to interpret directly: social media corpora are large, noisy, and weakly structured, with massive numbers of short posts collectively expressing a high-dimensional landscape of concerns, viewpoints, and events~\cite{dong2021review, cortis2021over, wang2017short}.

Taxonomies provide a useful abstraction for organizing such discourse. By arranging concepts into hierarchical structures, they enable coarse-to-fine exploration of themes, support interpretable analysis, and serve as a foundation for downstream tasks such as opinion mining, stance detection, and structured knowledge construction~\cite{liu2012automatic, cadilhac2010ontolexical, ju2022grape, wan2024tnt, kargupta2025beyond, pan2025taxonomy}. 
Taxonomy construction has broadly evolved from expert-curated hierarchies to scalable corpus-driven approaches based on clustering, hierarchical topic models, and term expansion, and more recently to LLM-assisted methods that can generate or refine taxonomies directly from text~\cite{griffiths2003hierarchical, shen2018hiexpan, shang2020nettaxo, lee2022taxocom, wan2024tnt, zhang2025llmtaxo, zhu2025context}. These advances have substantially reduced manual effort, but their assumptions remain only partially compatible with social media settings.

Constructing taxonomies from social media corpora presents three challenges that are not adequately addressed by existing methods. First, social media posts are short, informal, and semantically sparse \cite{qian2022co,zhao2021multi,zhao2023self}, often exhibiting lexical variability, sarcasm, and fragmented context, which makes it hard to obtain reliable semantic clusters~\cite{wang2017short, cortis2021over, dong2021review}. 
Second, social media streams are large and continuously growing, so methods that rely on repeated corpus-level passes can become computationally expensive and difficult to scale~\cite{wan2024tnt, kargupta2025taxoadapt}. 
Third, social media discourse is inherently dynamic: viewpoints evolve over time, and bursty real-world events can trigger short-lived yet meaningful concepts that static taxonomy induction may fail to preserve~\cite{deng2021cross, reveilhac2022systematic, zhu2025ratsd}. Existing taxonomy construction frameworks are largely developed for static corpora and therefore make limited use of temporal signals during hierarchy induction~\cite{lee2022taxocom, wan2024tnt, zhang2025llmtaxo, kargupta2025taxoadapt}.

\begin{figure}[t]
\centering
\includegraphics[width=\linewidth]{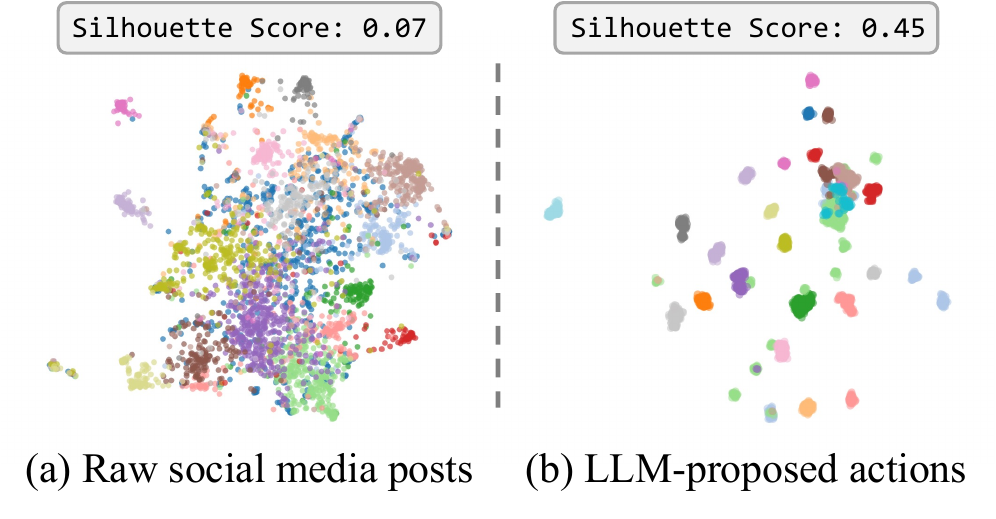}
\vspace{-15pt}
\caption{K-Means clustering of raw social media posts and their corresponding LLM-generated actions, where the number of clusters is selected by the highest silhouette score. Compared with raw posts, action representations form substantially clearer cluster structure, motivating action-level consolidation in EvoTaxo.}
\label{fig1}
\vspace{-10pt}
\end{figure}

To address these limitations, we propose \textbf{EvoTaxo}, a LLM-based framework for building and evolving taxonomies from social media streams. Instead of clustering raw posts directly, EvoTaxo converts temporally ordered posts into structured actions over the current taxonomy. As illustrated in Figure~\ref{fig1}, LLM-generated action representations exhibit substantially clearer cluster structure than raw social media posts, motivating action-level consolidation in place of direct clustering over noisy short texts. Structural actions are accumulated within time windows and consolidated using both semantic-only and time-aware views. A two-step review procedure then refines and arbitrates candidate actions before execution. In addition, each node maintains a \emph{concept memory bank} that stores its definition and conceptual boundaries, providing a persistent semantic anchor for subsequent taxonomy decisions and refinement. 

These design choices allow EvoTaxo to address the three challenges above in a unified way. By transforming posts into structured actions and grounding nodes with concept memory banks, the framework improves robustness to informal and underspecified language. By consolidating evidence incrementally rather than repeatedly traversing the full corpus, it improves scalability. By processing posts chronologically and incorporating time-aware consolidation, it captures discourse evolution and surfaces transient but meaningful concepts that static approaches may overlook. Experiments show that EvoTaxo produces more balanced taxonomies with lower leaf-assignment entropy, stronger corpus coverage, and consistently better structural quality than baselines. A case study on Reddit community \texttt{/r/ICE\_Raids} further shows that EvoTaxo tracks major shifts in discourse.

Our main contributions are as follows:
\begin{itemize}[noitemsep, leftmargin=*]
    \item \textbf{Problem Setting}: We propose EvoTaxo, to the best of our knowledge, the first LLM-based taxonomy construction framework tailored to temporally evolving social media corpora.
    \item \textbf{Methodology}: We formulate taxonomy construction as incremental structured editing over an evolving hierarchy, with targeted designs for robustness, scalability, and temporal sensitivity.
    \item \textbf{Empirical Findings}: We show empirically that EvoTaxo produces higher-quality taxonomies than strong baselines, while revealing interpretable discourse evolution over time.
\end{itemize}

\section{Related Work}
Automatic taxonomy construction has traditionally been studied in corpus-driven settings, where hierarchical structures are induced from text using unsupervised techniques such as topic modeling, clustering, and term expansion~\cite{zhang2018taxogen, shen2018hiexpan, shang2020nettaxo, lee2022taxocom}. 
While effective on relatively clean and semantically coherent text collections, these methods depend heavily on the quality of corpus-level groupings, which makes them less reliable for noisy and short social media texts.

More recently, LLMs \cite{ye2025llms4all} have been used to automate taxonomy construction. Some approaches combine clustering with LLM-based interpretation or hierarchy refinement~\cite{katz2024knowledge, zhang2025llmtaxo, zhu2025context}, while others use LLMs to induce an initial taxonomy and then iteratively refine or adapt it to a corpus~\cite{wan2024tnt, hsu2024chime, kargupta2025taxoadapt}. Although these methods reduce human effort, they remain limited for social media settings: clustering-based variants are still sensitive to corpus noise, and iterative LLM-based variants often require repeated corpus-level traversals for optimization.
In summary, existing taxonomy construction methods still lack a framework that is simultaneously robust to noisy social media text, computationally scalable on large corpora, and capable of exploiting temporal information during taxonomy induction. This gap motivates the design of EvoTaxo.


\section{Methodology}


\subsection{Preliminary}

\paragraph{Taxonomy and Problem Definition.}
Let $\mathcal{D}=\{(x_i,\tau_i)\}_{i=1}^{N}$ denote a temporally ordered social media corpus, where $x_i$ is a post and $\tau_i$ is its timestamp. We aim to incrementally construct a taxonomy over this stream.

We define a taxonomy as a rooted tree $\mathcal{T}=(\mathcal{V},\mathcal{E},r,\ell)$, where $\mathcal{V}$ is the set of nodes, $\mathcal{E}$ is the set of parent-child edges, $r$ is the root node, and $\ell(v)$ is the label of node $v$. In our setting, children of the root are \emph{topic} nodes, and children of topic nodes are \emph{subtopic} nodes. Given $\mathcal{D}$ and the root node $r$, EvoTaxo processes posts chronologically and incrementally updates $\mathcal{T}$ to organize the evolving corpus into coherent topics and subtopics.

\paragraph{Concept Memory Bank.}
A central difficulty in social media taxonomy construction is that node labels alone are often too underspecified to support stable decisions, since social media texts are short, informal, and semantically entangled. In particular, superficially similar wording may correspond to different concepts, while lexically different posts may express the same underlying concern. To mitigate this issue, we associate each node $v \in \mathcal{V}$ with a \emph{concept memory bank} (CMB), denoted by
\begin{align}
m(v) = \big(\delta(v),\, c^{+}(v),\, c^{-}(v)\big),
\end{align}
where $\delta(v)$ is a short definition of the concept represented by $v$, $c^{+}(v)$ is a set of inclusion cues, and $c^{-}(v)$ is a set of exclusion cues. The concept memory bank provides a more explicit semantic specification of the node, which helps preserve clearer concept boundaries and supports more consistent taxonomy refinement over time.

\subsection{Overview}

\begin{figure*}[t!]
\centering
\includegraphics[width=\textwidth]{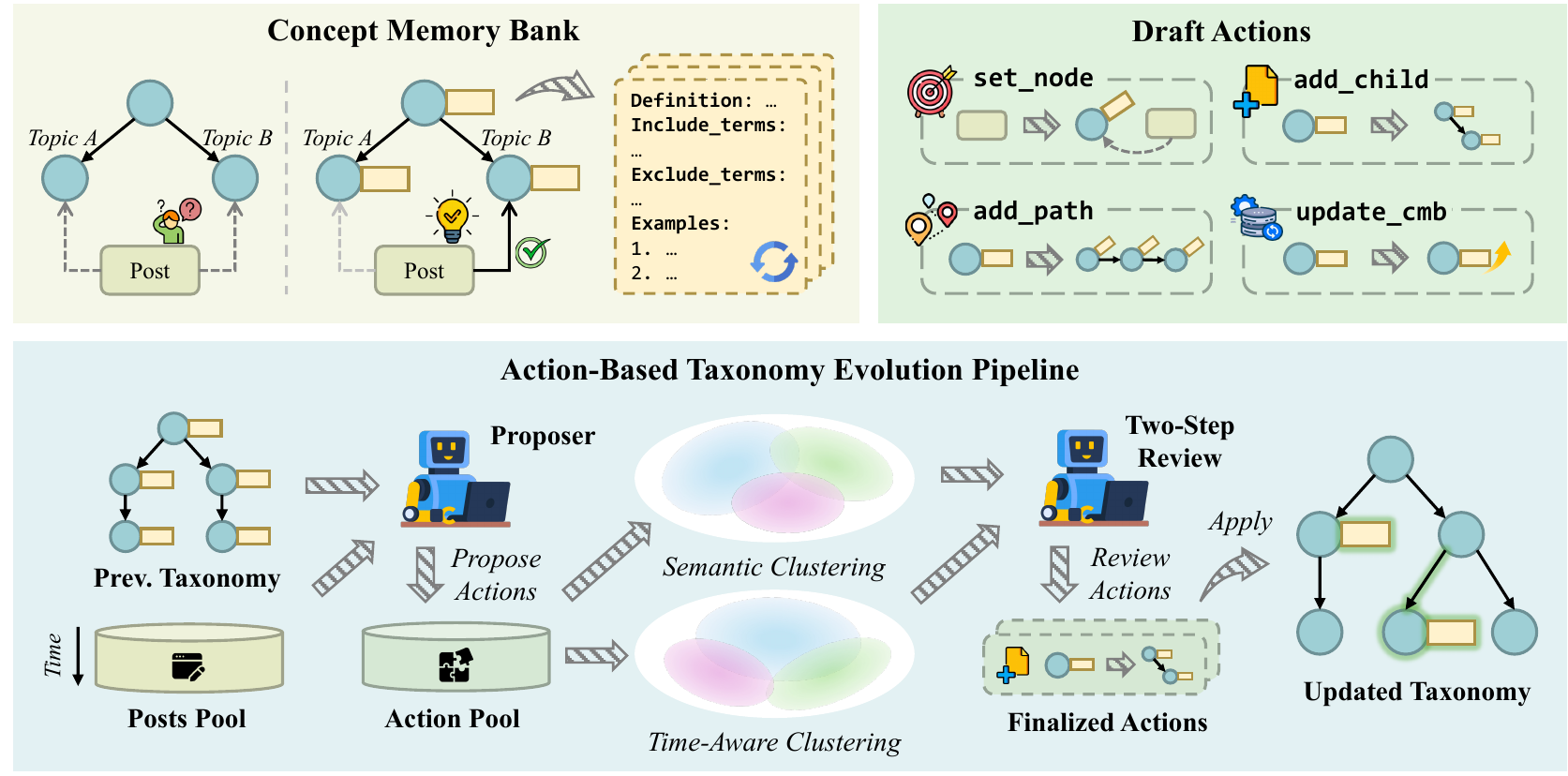}
\vspace{-10pt}
\caption{Overview of EvoTaxo. The framework starts from an LLM-generated seed taxonomy with concept memory banks and processes posts chronologically. Each post is mapped to a draft action. Structural actions are accumulated and consolidated at the window boundary under semantic and semantic-temporal views, then refined and arbitrated into final actions. These final actions update both the taxonomy and its temporal grounding records.}
\label{fig2}
\vspace{-10pt}
\end{figure*}

Figure~\ref{fig2} illustrates the full framework. EvoTaxo begins with an LLM-generated seed taxonomy given root node $r$. It then processes posts in chronological order and partitions the stream into a sequence of time windows $\mathcal{W}=\{W_1,\dots,W_M\}$. For each post $x_i$, the system generates a draft action conditioned on the current taxonomy and the concept memory banks of its nodes. Non-structural actions are executed immediately, whereas structural actions are accumulated in a backlog $\mathcal{B}$. At the boundary of each window $W_k$, EvoTaxo consolidates the current backlog under two complementary views, one based on semantic similarity and the other combining semantic similarity with temporal locality. The resulting candidate clusters are then examined by a two-step review procedure that first refines local evidence and then selects a globally compatible set of edits. Finally, the accepted actions are applied to update both the taxonomy structure and its temporal grounding records.

\subsection{Draft Action Generation} 

The first design question is how to convert noisy individual posts into a form that can contribute to taxonomy evolution. Since a single post rarely contains enough evidence to justify a structural change on its own, EvoTaxo first transforms each post into a draft action. This representation makes later aggregation more reliable since it avoids clustering directly over raw and noisy social media texts.

Let $\mathcal{T}_i$ denote the current taxonomy state when processing post $x_i$ within time window $W_k$, including both the tree structure and the CMB of each node. The proposer is an LLM-based function
\begin{align}
a_i = f_{\mathrm{prop}}(x_i,\mathcal{T}_i),
\end{align}
which maps post $x_i$ to a draft action $a_i$. Each action is associated with the timestamp of its source post, denoted by $\tau(a_i)=\tau_i$.

EvoTaxo considers five draft action types:

\begin{itemize}[noitemsep, leftmargin=*]
    \item \textbf{\texttt{set\_node}}: assign the post directly to an existing topic or subtopic node;
    \item \textbf{\texttt{add\_child}}: create a new child node under an existing node;
    \item \textbf{\texttt{add\_path}}: create a short path of the form root $\rightarrow$ topic $\rightarrow$ subtopic;
    \item \textbf{\texttt{update\_cmb}}: revise the concept memory bank of an existing node;
    \item \textbf{\texttt{skip\_post}}: mark the post as uninformative for taxonomy construction.
\end{itemize}

In particular, \texttt{set\_node} and \texttt{skip\_post} are executed immediately since they do not modify the taxonomy structure. By contrast, structural actions, namely \texttt{add\_child}, \texttt{add\_path}, and \texttt{update\_cmb}, are stored in a backlog $\mathcal{B}$ for later aggregation. 

By postponing structural execution until the window boundary, EvoTaxo reduces the risk that the taxonomy overreacts to isolated posts, idiosyncratic phrasing, or transient noise.

\subsection{Dual-View Action Consolidation}

The next challenge is how to consolidate the draft actions in the backlog. Purely semantic clustering may smooth away short-lived but meaningful discourse patterns. Therefore, EvoTaxo consolidates actions under two complementary views so that persistent semantic patterns and temporally concentrated discourse bursts can both be preserved.

Given the backlog $\mathcal{B}$ at a window boundary, we first partition it into smaller buckets $\mathcal{G}$, where each bucket is indexed by the action's type and its target node. This partition prevents actions with different edit semantics from being clustered together.

Within each bucket $\mathcal{G}$, each draft action $a \in \mathcal{G}$ is converted into a semantic embedding vector $z(a)$ based on its action content. We then cluster actions in two passes.

The first pass captures semantic consistency across time. For two actions $a_m$ and $a_n$, we define the semantic distance as
\begin{align}
d_{\mathrm{sem}}(a_m,a_n)=1-\cos\big(z(a_m),z(a_n)\big).
\end{align}
This view favors action groups that express similar structural update intent across time.

The second pass additionally considers temporal locality. This is important because some meaningful concepts appear as concentrated bursts tied to unfolding events. Specifically, we define the normalized temporal distance within bucket $\mathcal{G}$ as
\begin{align}
d_{\mathrm{time}}(a_m,a_n)
=
\frac{|\tau(a_m)-\tau(a_n)|}
{\tau_{\max}(\mathcal{G})-\tau_{\min}(\mathcal{G})+\epsilon},
\end{align}
where $\epsilon$ prevents division by zero. We then combine semantic and temporal information through
\begin{align}
d_{\mathrm{joint}}(a_m,a_n)
=
(1-\lambda)\, d_{\mathrm{sem}}(a_m,a_n)
+ \notag\\
\lambda\, d_{\mathrm{time}}(a_m,a_n),
\end{align}
where $\lambda \in [0,1]$ controls the contribution of temporal locality.

For both views, we apply HDBSCAN~\cite{mcinnes2017hdbscan} as the clustering algorithm, since it does not require a predefined number of clusters and allows us to define custom distance functions.

The semantic pass highlights globally recurring structural intents, whereas the joint pass highlights semantically coherent actions that are also concentrated in time. EvoTaxo retains candidate clusters from both of them for downstream review.

\subsection{Action Refinement and Arbitration}

After consolidation, EvoTaxo must decide which candidate clusters should actually change the taxonomy. This requires more than simply accepting all clusters, because some clusters may still be noisy, redundant, or mutually incompatible. For this reason, EvoTaxo uses a two-step decision procedure that first refines local cluster evidence and then resolves global conflicts among candidate edits.

\paragraph{Cluster-level refinement.}
The first step operates at the level of an individual cluster. Its purpose is to determine whether the actions within a cluster support a coherent structural update. For each cluster, an LLM reviewer is given representative posts, the corresponding draft actions, and the current taxonomy context. Based on this evidence, the reviewer either defers the cluster or rewrites it into one of more \emph{refined actions}. A refined action is therefore a cluster-level abstraction that summarizes what the supporting posts collectively justify. 

\paragraph{Window-level arbitration.}
Although refined actions are cleaner than raw draft actions, they are still produced independently from different clusters. Consequently, the current window may contain overlapping additions, competing node names, or edits that cannot be applied together without reducing structural coherence. To address this issue, EvoTaxo performs a second review step over the entire set of refined actions generated at the current window boundary. An LLM arbitration reviewer jointly examines these refined actions and outputs a set of \emph{final actions} to be executed. 

After arbitration, committed proposals are removed from the backlog, while deferred or unsupported evidence is retained for possible reconsideration in future windows.

\subsection{Taxonomy Update and Grounding}

Once the final actions for a window are determined, EvoTaxo applies them deterministically to update the taxonomy. When a final action creates a new node or path, the corresponding nodes and edges are added to $\mathcal{T}$. When a final action updates a concept memory bank, the memory bank of the corresponding node is revised accordingly. In addition, associated posts are grounded to the node implied by the executed action. This grounding links the taxonomy back to the supporting corpus evidence and improves interpretability.
Appendix~\ref{app:pseudo_code} provides the pseudo code for the full EvoTaxo pipeline.


\section{Experiments}
\label{exp}

\subsection{Experiment Setup}
\label{sec:exp_setup}

\paragraph{Datasets.}
We evaluate EvoTaxo on datasets collected from two Reddit communities: \texttt{/r/opiates} and \texttt{/r/ICE\_Raids}. The two datasets stress complementary challenges of social media taxonomy induction. \texttt{/r/opiates} is a high-noise, experience-centered community whose posts often mix symptoms, usage practices, recovery narratives, advice seeking, and emotionally charged personal accounts, making it difficult to derive a stable hierarchical structure from informal and semantically entangled language. In contrast, \texttt{/r/ICE\_Raids} is an event-driven corpus with strong temporal dynamics, where discussions are shaped by unfolding incidents, rumor diffusion, political reactions, and short-lived news cycles. This makes it a useful test bed for evaluating whether temporal signals help identify bursty yet meaningful concepts.

We use two temporal settings for \texttt{/r/opiates}: 2015--2024 with yearly windows (\textit{N} = 15,846 posts), and 2022--2024 with quarterly windows (\textit{N} = 8,582 posts). For \texttt{/r/ICE\_Raids}, we use data from February 2025 to February 2026 with monthly windows (\textit{N} = 18,394 posts). Data preprocessing details are deferred to Appendix~\ref{app:data}.
\begin{table*}[t]
\centering
\small
\caption{Main results across datasets. EvoTaxo achieves the best trade-off between assignment quality and structural quality while maintaining a relatively balanced taxonomy. Tokens are reported in millions (M).}
\label{tab:main_results}
\begin{tabular}{ll|cc|cccccc|r}
\toprule
Dataset & Method 
& Nodes 
& Leaf
& Ent$\downarrow$ 
& Uncls$\downarrow$ 
& NLIV-S$\uparrow$ 
& Path$\uparrow$ 
& Sib-C$\uparrow$ 
& Sib-S$\uparrow$ 
& Tokens \\
\midrule
\multirow{5}{*}{\shortstack{/r/Opiates \\ (2015-2024)}}
& KN
& 7 & 7 
& 0.93 & 0.85 & 0.40 & 3.16 & 4.00 & 3.00 & <1\\
& Chain-of-Layer 
& 40 & 40 
& \underline{0.83} & \textbf{0.16} & \underline{0.58} & \underline{4.13} & 4.00 & 3.00 & <1\\
& TnT-LLM 
& 25 & 25 
& 0.90 & 0.29 & 0.36 & 3.04 & 4.00 & 3.00 & 60.9 \\
& TaxoAdapt 
& 73 & 62 
& 0.95 & 0.30 & 0.52 & 3.38 & \underline{4.27} & \underline{3.36} & 164.4 \\
& \textbf{EvoTaxo (Ours)}
& 25 & 19 
& \textbf{0.79} & \underline{0.22} & \textbf{0.83} & \textbf{4.68} & \textbf{4.60} & \textbf{3.40} & 35.3 \\
\midrule
\multirow{5}{*}{\shortstack{/r/Opiates \\ (2022-2024)}}
& KN
& 6 & 6 
& 0.93 & 0.86 & 0.46 & 2.83 & 3.00 & \textbf{4.00} & <1\\
& Chain-of-Layer 
& 43 & 43 
& \underline{0.85} & \textbf{0.17} & \underline{0.58} & 3.85 & 2.50 & 3.00 & <1\\
& TnT-LLM 
& 25 & 25 
& 0.91 & 0.28 & 0.35 & 3.50 & \underline{4.00} & 3.50 & 33.5 \\
& TaxoAdapt 
& 72 & 63 
& 0.94 & 0.30 & 0.55 & \underline{4.00} & 3.67 & \underline{3.75} & 92.1 \\
& \textbf{EvoTaxo (Ours)} 
& 29 & 20 
& \textbf{0.80} & \underline{0.22} & \textbf{0.80} & \textbf{4.54} & \textbf{4.43} & \underline{3.75} & 18.3 \\
\midrule
\multirow{5}{*}{/r/ICE\_Raids}
& KN
& 5 & 5 
& 0.89 & 0.36 & \underline{0.58} & 3.60 & 4.00 & \textbf{4.00} & <1\\
& Chain-of-Layer 
& 33 & 33
& \underline{0.77} & 0.24 & 0.24 & 4.00 & 4.00 & 3.00 & <1\\
& TnT-LLM 
& 10 & 10 
& 0.86 & 0.26 & 0.53 & 4.10 & 4.00 & 3.00 & 65.7 \\
& TaxoAdapt 
& 71 & 60 
& 0.93 & \textbf{0.20} & 0.18 & \underline{4.67} & \underline{4.46} & 3.36 & 166.3 \\
& \textbf{EvoTaxo (Ours)} 
&  32 & 27 
& \textbf{0.74} & \underline{0.23} & \textbf{0.72} & \textbf{4.85} & \textbf{4.80} & \underline{3.40} & 37.8 \\
\bottomrule
\end{tabular}
\end{table*}
\begin{table*}[t]
\centering
\small
\caption{Ablation study on \texttt{/r/opiates} (2015--2024). Removing CMB weakens semantic stability, removing time-aware clustering harms coverage, and removing arbitration leads to reduced hierarchy consistency.}
\label{tab:ablation}
\setlength{\tabcolsep}{9pt}
\begin{tabular}{l|cc|cccccc}
\toprule
Method 
& Nodes 
& Leaf
& Ent$\downarrow$ 
& Uncls$\downarrow$ 
& NLIV-S$\uparrow$ 
& Path$\uparrow$ 
& Sib-C$\uparrow$ 
& Sib-S$\uparrow$ \\
\midrule
EvoTaxo
& 25 & 19 
& \textbf{0.79} & \underline{0.22} & \textbf{0.83} & \textbf{4.68} & \textbf{4.60} & 3.40 \\
- w/o CMB
& 14 & 8
& 0.82 & 0.28 & 0.78 & \underline{4.53} & \underline{4.50} & \underline{3.50} \\
- w/o Time-Aware Clustering
& 10 & 8
& \underline{0.80} & 0.50 & \underline{0.80} & 4.50 & 4.00 & \textbf{4.00} \\
- w/o Arbitration
& 70 & 54
& 0.92 & \textbf{0.17} & 0.74 & 4.00 & 3.67 & 3.00 \\

\bottomrule
\end{tabular}
\end{table*}

\paragraph{Baselines.}
We compare EvoTaxo against representative clustering-incorporated and pure LLM-based taxonomy construction baselines, including \textbf{Knowledge Navigator} (KN)~\cite{katz2024knowledge}, \textbf{Chain-of-Layer}~\cite{zeng2024chain}, \textbf{TnT-LLM}~\cite{wan2024tnt}, and \textbf{TaxoAdapt}~\cite{kargupta2025taxoadapt}. 
We use \texttt{GPT-4o-mini}~\cite{hurst2024gpt} as the LLM backbone for all the baselines and EvoTaxo. More experiment and baseline details are discussed in Appendix~\ref{app:exp_details}.

\paragraph{Evaluation Metrics.}
We assess EvoTaxo from two complementary perspectives: (1) post-level assignment quality, which measures how decisively posts can be mapped to taxonomy leaves, and (2) structural quality, which measures whether the learned hierarchy is semantically well-formed.

\begin{itemize}[noitemsep, leftmargin=*]
    \item \textbf{Leaf Assignment Entropy.} For each post, we use an NLI model to score all leaf labels together with an additional \texttt{others} class. After removing \texttt{others}, we renormalize the distribution and compute the normalized entropy
    \begin{align}
    H(x) = - \frac{1}{\log |\mathcal{L}|} \sum_{\ell \in \mathcal{L}} p(\ell \mid x)\log p(\ell \mid x),
    \end{align}
    where $\mathcal{L}$ is the set of leaf nodes. Lower values indicate sharper post-to-leaf assignment.

    \item \textbf{Unclassified Rate.} The fraction of posts whose top predicted label is \texttt{others}. Lower values suggest more posts are covered by actual leaf nodes in the taxonomy.

    \item \textbf{NLIV-S}~\cite{wullschleger2025no}. For each parent--child edge, we use an NLI model to measure whether the child can be interpreted as a more specific abstraction of the parent. Higher values indicate more semantically valid hierarchical relations.

    \item \textbf{Path Granularity}~\cite{kargupta2025taxoadapt}. An LLM judge evaluates whether each root-to-leaf path becomes progressively more specific in a coherent and sensible way. 

    \item \textbf{Sibling Coherence}~\cite{kargupta2025taxoadapt}. An LLM judge evaluates whether the children under the same parent belong together conceptually and are of a comparable level of abstraction scope. 

    \item \textbf{Sibling Separability.} An LLM judge evaluates whether sibling nodes under the same parent are clearly distinguishable from one another, rather than overlapping or redundant. 
\end{itemize}


We use \texttt{facebook/bart-large-mnli}\footnote{https://huggingface.co/facebook/bart-large-mnli} as the NLI model and \texttt{GPT-4o-mini} as the LLM judge. LLM-based metrics are scored on a 5-point scale~\cite{li2026grading}. We provide the prompts for evaluation in Appendix~\ref{app:eval_prompts} and the LLM-human agreement analysis in Appendix~\ref{app:agreement}.

\subsection{Main Results}

\begin{figure*}[t!]
    \centering
    \includegraphics[width=\textwidth]{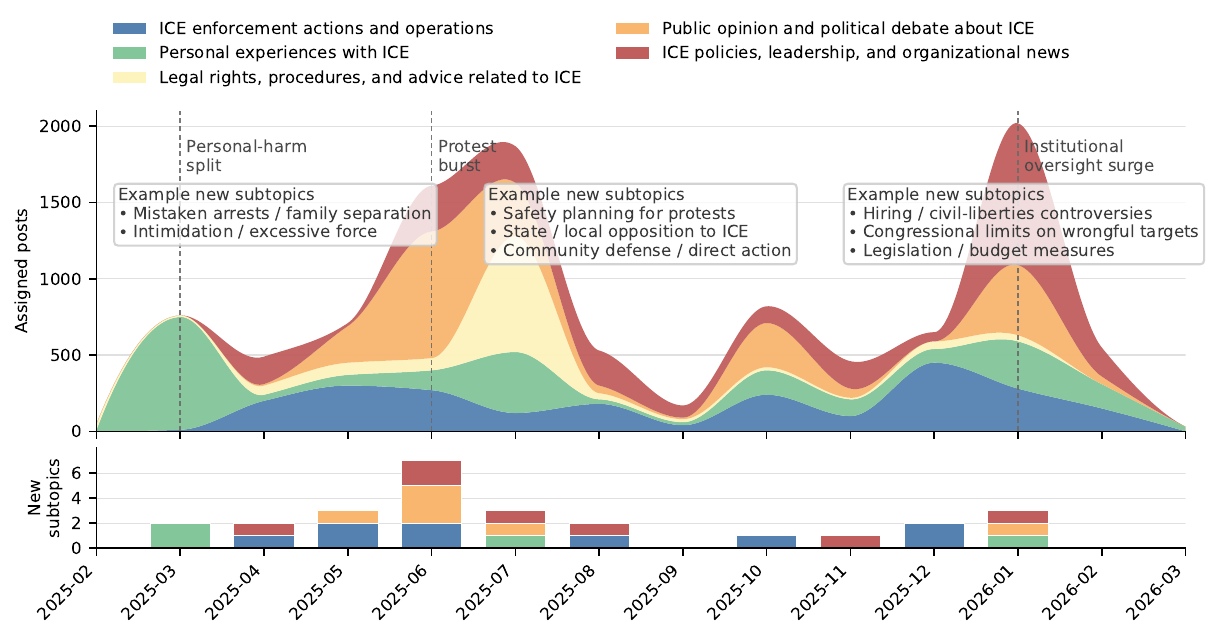}
    \vspace{-8pt}
    \caption{Monthly evolution of the taxonomy induced from \texttt{/r/ICE\_Raids}, aggregated to the five top-level topics. The figure highlights three major turning points: an early concentration on personal-harm narratives, a mid-period protest-driven burst, and a later shift toward institutional oversight and policy discussion. The lower panel shows the number of newly introduced subtopics per month.}
    \label{fig:ice_trend_example}
    \vspace{-8pt}
\end{figure*}

\begin{figure*}[t]
    \centering
    \includegraphics[width=\textwidth]{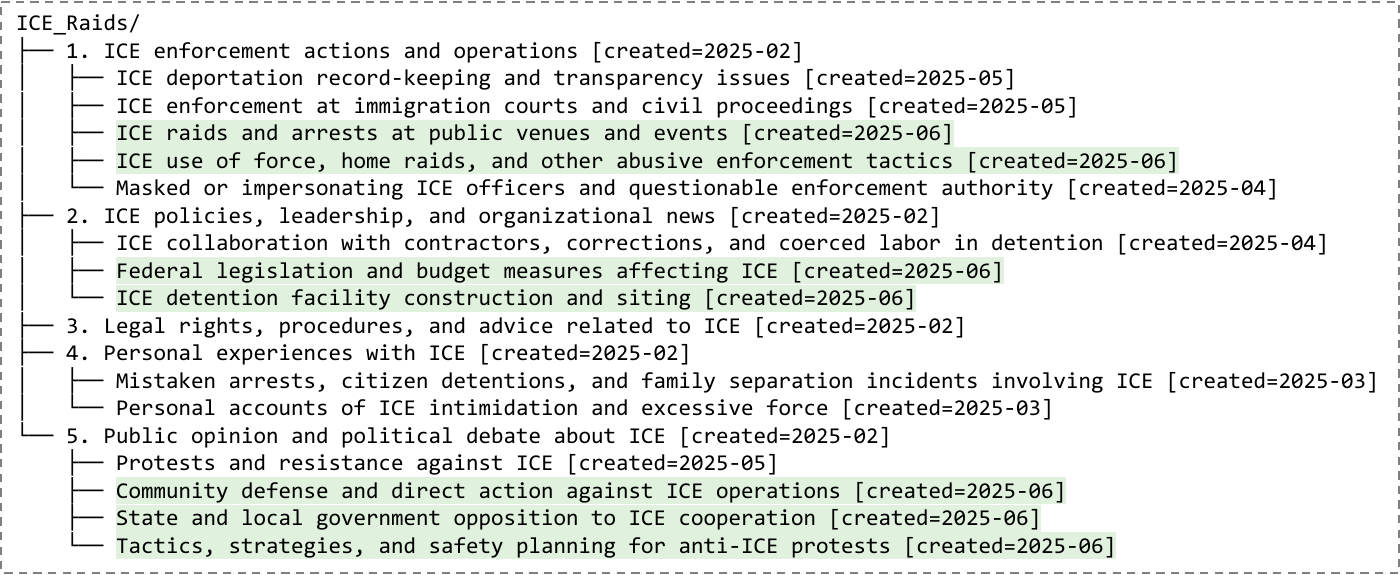}
    \vspace{-8pt}
    \caption{Taxonomy snapshot for \texttt{/r/ICE\_Raids} after the June 2025 update. Green highlights mark subtopics newly added in June, illustrating how EvoTaxo converts a temporally concentrated protest-driven discourse burst into explicit taxonomy growth.}
    \label{fig:ice_june_snapshot}
    \vspace{-8pt}
\end{figure*}

Table~\ref{tab:main_results} reports the main results across the three evaluation settings. 

On the post-level metrics, EvoTaxo consistently attains the lowest \textit{Leaf Assignment Entropy}, indicating that its leaf categories provide clearer destinations for individual posts. It also maintains a low \textit{Unclassified Rate} and clearly outperforms baselines with taxonomy sizes closer to its own. These results suggest that EvoTaxo induces taxonomies whose leaves are both sufficiently specific for confident assignment and sufficiently representative to cover a large proportion of the corpus.

EvoTaxo also performs strongest on structure-oriented metrics. It consistently achieves the best \textit{NLIV-S}, \textit{Path Granularity}, and \textit{Sibling Coherence}, indicating that its parent--child relations are more semantically valid and that its paths exhibit cleaner broad-to-specific refinement. By contrast, KN, Chain-of-Layer, and TnT-LLM produce only one-layer taxonomies, which limits their ability to express finer semantic stratification even when their sibling sets remain relatively separable.

Moreover, the token statistics show that EvoTaxo is also more efficient than the other post-level LLM baselines, i.e., TnT-LLM and TaxoAdapt. Across all settings, EvoTaxo uses substantially fewer tokens while achieving stronger overall quality. This efficiency gain is consistent with the design of the framework: EvoTaxo processes the corpus in a single pass and accumulates structural evidence incrementally, rather than repeatedly traversing the corpus for iterative taxonomy refinement. 

Overall, EvoTaxo achieves the strongest balance between post-level assignment quality, taxonomy structural quality, and computational efficiency, while maintaining a relatively compact taxonomy.

\subsection{Ablation Study.}

Table~\ref{tab:ablation} reports an ablation study on \texttt{/r/opiates} (2015--2024). Each component contributes to taxonomy quality in a distinct way.

Removing the \emph{concept memory bank} (CMB) substantially reduces the number of nodes and leaves, while increasing both \textit{Leaf Assignment Entropy} and \textit{Unclassified Rate}. This suggests that, without CMB, the system is less able to preserve and accumulate fine-grained concept distinctions. As a result, the taxonomy collapses into a smaller set of broader categories, leaving more posts insufficiently covered by leaf nodes. CMB therefore plays an important role in maintaining semantic boundaries and supporting stable fine-grained expansion.

Removing \emph{time-aware clustering} further shrinks the taxonomy and causes the largest increase in \textit{Unclassified Rate}, indicating a substantial loss of corpus coverage. The resulting hierarchy becomes much smaller and less expressive, suggesting that semantic-only consolidation is less able to surface temporally localized meaningful distinctions. This result supports the role of time-aware clustering in preserving emerging concepts that are coherent not only semantically but also temporally.

Removing \emph{arbitration review} has the opposite effect on taxonomy size: the number of nodes and leaves increases dramatically, showing that many overlapping or conflicting refined actions are applied simultaneously. However, this larger taxonomy does not improve quality. Instead, entropy increases, \textit{NLIV-S} decreases, and \textit{Sibling Coherence} drops noticeably. This indicates that arbitration is crucial for resolving redundancy and conflict before execution; without it, the taxonomy grows more aggressively but becomes less selective and internally less consistent.

Overall, the ablation results confirm the intended roles of the three components: CMB improves semantic stability, time-aware clustering improves temporal sensitivity and corpus coverage, and arbitration prevents uncontrolled structural expansion while preserving hierarchy quality.

\subsection{Case Study}

To examine whether EvoTaxo captures temporal discourse structure rather than merely producing a static hierarchy, we analyze the monthly evolution of the taxonomy induced from \texttt{/r/ICE\_Raids}. Figure~\ref{fig:ice_trend_example} aggregates post assignments at the top-topic level and reveals several clear shifts over time.

The first salient transition occurs in March 2025, when the discussion is dominated by \emph{Personal experiences with ICE}. EvoTaxo introduces subtopics centered on mistaken arrests, family separation, intimidation, and excessive force. This suggests that the taxonomy initially organizes around recurrent first-hand harm narratives rather than prematurely fragmenting broader institutional themes.

A second major shift appears in June 2025, when discourse moves sharply toward \emph{Public opinion and political debate about ICE}. During this period, EvoTaxo adds protest-related subtopics such as safety planning, state and local opposition, and community defense. These additions are both semantically coherent and temporally concentrated, illustrating the value of time-aware consolidation for preserving bursty but meaningful discourse phases that static induction could easily absorb into broader political categories. Figure~\ref{fig:ice_june_snapshot} provides a snapshot of the taxonomy after the June 2025 update and shows that these new branches are introduced in a targeted rather than diffuse manner, concentrated around the parts of the hierarchy most directly affected by the protest surge.

A final transition emerges around January 2026, when the dominant emphasis shifts toward \emph{ICE policies, leadership, and organizational news}. Newly added subtopics focus on hiring controversies, civil-liberties oversight, and congressional limits on wrongful targets, indicating that the center of discourse has moved from street-level events and mobilization toward institutional accountability and governance.

Overall, the case study shows that EvoTaxo tracks discourse evolution at both the structural and semantic levels: stable concepts are retained, while temporally localized developments are surfaced as new branches when they become salient. Appendix~\ref{app:ice_microbursts} further examines lower-volume but sharply time-concentrated subtopics, which provide more direct evidence for the utility of the temporal clustering view.

\section{Conclusion}

We introduced \textbf{EvoTaxo}, a time-aware LLM-based framework for taxonomy construction from social media streams. By formulating taxonomy induction as incremental structural editing over temporally ordered posts, EvoTaxo improves robustness to noisy short texts, supports efficient evidence accumulation, and captures temporally localized discourse shifts. Experiments on two Reddit corpora show that EvoTaxo produces more balanced taxonomies than baselines, with clearer post-to-leaf assignment, strong coverage, and better structural quality at comparable taxonomy size. The case study further shows that EvoTaxo can reveal how discourse evolves over time by introducing new branches when emerging concepts become salient.

\section{Limitations}

EvoTaxo has several limitations. First, the framework depends on the quality of LLM-generated actions and reviews. Although the staged design helps reduce the impact of individual errors, inaccurate or biased generations can still affect structural updates, especially in low-support regions of the taxonomy. 
Second, some of our evaluation metrics rely on NLI models and LLM judges rather than fully human annotation. These automatic judges provide scalable proxies for taxonomy quality, but they may not perfectly align with expert judgment in all cases.
Third, EvoTaxo requires the choice of a temporal window size, and the optimal granularity may vary across domains. Coarser windows may smooth over bursty developments, whereas finer windows may increase fragmentation and reduce the amount of evidence available for consolidation. Fourth, our experiments focus on two Reddit communities with distinct but still platform-specific discourse characteristics. Although these settings cover both noisy experiential discussion and event-driven discussion, further validation is needed on other platforms, languages, and domains.

\section{Ethical Considerations}

We use publicly available Reddit data from communities discussing sensitive topics, including drug use and immigration enforcement. Although these data are public, they involve human-generated content that may still carry privacy and re-identification risks. Accordingly, we focus on aggregate taxonomy construction rather than individual-level analysis, avoid identity inference, and minimize reproduction of sensitive text in qualitative discussion.

EvoTaxo is intended as a research tool for analyzing evolving discourse, not for surveillance, profiling, or enforcement. The induced taxonomies should be interpreted as heuristic summaries rather than ground truth, since they may reflect both dataset biases and biases introduced by LLM and NLI components. Future work should further examine representational harms, bias mitigation, and appropriate human oversight in sensitive application settings.

\bibliography{custom}

@article{ye2025llms4all,
  title={Llms4all: A review of large language models across academic disciplines},
  author={Ye, Yanfang and Zhang, Zheyuan and Ma, Tianyi and Wang, Zehong and Li, Yiyang and Hou, Shifu and Sun, Weixiang and Shi, Kaiwen and Ma, Yijun and Song, Wei and others},
  journal={arXiv preprint arXiv:2509.19580},
  year={2025}
}

@inproceedings{ju2022grape,
  title={Grape: Knowledge graph enhanced passage reader for open-domain question answering},
  author={Ju, Mingxuan and Yu, Wenhao and Zhao, Tong and Zhang, Chuxu and Ye, Yanfang},
  booktitle={Findings of the Association for Computational Linguistics: EMNLP 2022},
  pages={169--181},
  year={2022}
}

@article{qian2022co,
  title={Co-modality graph contrastive learning for imbalanced node classification},
  author={Qian, Yiyue and Zhang, Chunhui and Zhang, Yiming and Wen, Qianlong and Ye, Yanfang and Zhang, Chuxu},
  journal={Advances in Neural Information Processing Systems},
  volume={35},
  pages={15862--15874},
  year={2022}
}

@inproceedings{zhao2023self,
  title={Self-supervised graph structure refinement for graph neural networks},
  author={Zhao, Jianan and Wen, Qianlong and Ju, Mingxuan and Zhang, Chuxu and Ye, Yanfang},
  booktitle={Proceedings of the sixteenth ACM international conference on web search and data mining},
  pages={159--167},
  year={2023}
}

@inproceedings{zhao2021multi,
  title={Multi-view self-supervised heterogeneous graph embedding},
  author={Zhao, Jianan and Wen, Qianlong and Sun, Shiyu and Ye, Yanfang and Zhang, Chuxu},
  booktitle={Joint European conference on machine learning and knowledge discovery in databases},
  pages={319--334},
  year={2021},
  organization={Springer}
}

@article{lazer2009computational,
  title={Computational social science},
  author={Lazer, David and Pentland, Alex and Adamic, Lada and Aral, Sinan and Barab{\'a}si, Albert-L{\'a}szl{\'o} and Brewer, Devon and Christakis, Nicholas and Contractor, Noshir and Fowler, James and Gutmann, Myron and others},
  journal={Science},
  volume={323},
  number={5915},
  pages={721--723},
  year={2009},
  publisher={American Association for the Advancement of Science}
}

@article{zhuravskaya2020political,
  title={Political effects of the internet and social media},
  author={Zhuravskaya, Ekaterina and Petrova, Maria and Enikolopov, Ruben},
  journal={Annual review of economics},
  volume={12},
  number={1},
  pages={415--438},
  year={2020},
  publisher={Annual Reviews}
}

@article{dong2021review,
  title={A review of social media-based public opinion analyses: Challenges and recommendations},
  author={Dong, Xuefan and Lian, Ying},
  journal={Technology in Society},
  volume={67},
  pages={101724},
  year={2021},
  publisher={Elsevier}
}

@article{deng2021cross,
  title={Cross-platform comparative study of public concern on social media during the COVID-19 pandemic: An empirical study based on Twitter and Weibo},
  author={Deng, Wen and Yang, Yi},
  journal={International Journal of Environmental Research and Public Health},
  volume={18},
  number={12},
  pages={6487},
  year={2021},
  publisher={MDPI}
}

@inproceedings{li2024pro,
  title={Pro-woman, anti-man? identifying gender bias in stance detection},
  author={Li, Yingjie and Zhang, Yue},
  booktitle={Findings of the Association for Computational Linguistics: ACL 2024},
  pages={3229--3236},
  year={2024}
}

@inproceedings{zhu2025ratsd,
  title={RATSD: Retrieval augmented truthfulness stance detection from social media posts toward factual claims},
  author={Zhu, Zhengyuan and Zhang, Zeyu and Zhang, Haiqi and Li, Chengkai},
  booktitle={Findings of the Association for Computational Linguistics: NAACL 2025},
  pages={3366--3381},
  year={2025}
}

@article{reveilhac2022systematic,
  title={A systematic literature review of how and whether social media data can complement traditional survey data to study public opinion},
  author={Reveilhac, Maud and Steinmetz, Stephanie and Morselli, Davide},
  journal={Multimedia tools and applications},
  volume={81},
  number={7},
  pages={10107--10142},
  year={2022},
  publisher={Springer}
}

@article{cortis2021over,
  title={Over a decade of social opinion mining: a systematic review},
  author={Cortis, Keith and Davis, Brian},
  journal={Artificial intelligence review},
  volume={54},
  number={7},
  pages={4873--4965},
  year={2021},
  publisher={Springer}
}

@inproceedings{wang2017short,
  title={A short survey on taxonomy learning from text corpora: Issues, resources and recent advances},
  author={Wang, Chengyu and He, Xiaofeng and Zhou, Aoying},
  booktitle={Proceedings of the 2017 Conference on Empirical Methods in Natural Language Processing},
  pages={1190--1203},
  year={2017}
}

@inproceedings{liu2012automatic,
  title={Automatic taxonomy construction from keywords},
  author={Liu, Xueqing and Song, Yangqiu and Liu, Shixia and Wang, Haixun},
  booktitle={Proceedings of the 18th ACM SIGKDD international conference on Knowledge discovery and data mining},
  pages={1433--1441},
  year={2012}
}

@inproceedings{shen2018hiexpan,
  title={Hiexpan: Task-guided taxonomy construction by hierarchical tree expansion},
  author={Shen, Jiaming and Wu, Zeqiu and Lei, Dongming and Zhang, Chao and Ren, Xiang and Vanni, Michelle T and Sadler, Brian M and Han, Jiawei},
  booktitle={Proceedings of the 24th ACM SIGKDD International Conference on Knowledge Discovery \& Data Mining},
  pages={2180--2189},
  year={2018}
}

@inproceedings{lee2022taxocom,
  title={Taxocom: Topic taxonomy completion with hierarchical discovery of novel topic clusters},
  author={Lee, Dongha and Shen, Jiaming and Kang, SeongKu and Yoon, Susik and Han, Jiawei and Yu, Hwanjo},
  booktitle={Proceedings of the ACM Web Conference 2022},
  pages={2819--2829},
  year={2022}
}

@article{griffiths2003hierarchical,
  title={Hierarchical topic models and the nested Chinese restaurant process},
  author={Griffiths, Thomas and Jordan, Michael and Tenenbaum, Joshua and Blei, David},
  journal={Advances in neural information processing systems},
  volume={16},
  year={2003}
}

@inproceedings{zhang2025llmtaxo,
  title={LLMTaxo: Leveraging Large Language Models for Constructing Taxonomy of Factual Claims from Social Media},
  author={Zhang, Haiqi and Zhu, Zhengyuan and Zhang, Zeyu and Li, Chengkai},
  booktitle={Findings of the Association for Computational Linguistics: ACL 2025},
  pages={19627--19641},
  year={2025}
}

@inproceedings{kargupta2025beyond,
  title={Beyond true or false: Retrieval-augmented hierarchical analysis of nuanced claims},
  author={Kargupta, Priyanka and Tian, Runchu and Han, Jiawei},
  booktitle={Proceedings of the 63rd Annual Meeting of the Association for Computational Linguistics (Volume 1: Long Papers)},
  pages={29664--29679},
  year={2025}
}

@inproceedings{wan2024tnt,
  title={Tnt-llm: Text mining at scale with large language models},
  author={Wan, Mengting and Safavi, Tara and Jauhar, Sujay Kumar and Kim, Yujin and Counts, Scott and Neville, Jennifer and Suri, Siddharth and Shah, Chirag and White, Ryen W and Yang, Longqi and others},
  booktitle={Proceedings of the 30th ACM SIGKDD Conference on Knowledge Discovery and Data Mining},
  pages={5836--5847},
  year={2024}
}

@inproceedings{cadilhac2010ontolexical,
  title={Ontolexical resources for feature-based opinion mining: a case-study},
  author={Cadilhac, Ana{\"\i}s and Benamara, Farah and Aussenac-Gilles, Nathalie},
  booktitle={Proceedings of the 6th Workshop on Ontologies and Lexical Resources},
  pages={77--86},
  year={2010}
}

@inproceedings{pan2025taxonomy,
  title={Taxonomy-driven knowledge graph construction for domain-specific scientific applications},
  author={Pan, Huitong and Zhang, Qi and Adamu, Mustapha and Dragut, Eduard and Latecki, Longin Jan},
  booktitle={Findings of the Association for Computational Linguistics: ACL 2025},
  pages={4295--4320},
  year={2025}
}

@inproceedings{zhu2025context,
  title={Context-aware hierarchical taxonomy generation for scientific papers via llm-guided multi-aspect clustering},
  author={Zhu, Kun and Liao, Lizi and Gu, Yuxuan and Huang, Lei and Feng, Xiaocheng and Qin, Bing},
  booktitle={Proceedings of the 2025 Conference on Empirical Methods in Natural Language Processing},
  pages={15627--15645},
  year={2025}
}

@inproceedings{zhang2018taxogen,
  title={Taxogen: Unsupervised topic taxonomy construction by adaptive term embedding and clustering},
  author={Zhang, Chao and Tao, Fangbo and Chen, Xiusi and Shen, Jiaming and Jiang, Meng and Sadler, Brian and Vanni, Michelle and Han, Jiawei},
  booktitle={Proceedings of the 24th ACM SIGKDD international conference on knowledge discovery \& data mining},
  pages={2701--2709},
  year={2018}
}

@inproceedings{shang2020nettaxo,
  title={Nettaxo: Automated topic taxonomy construction from text-rich network},
  author={Shang, Jingbo and Zhang, Xinyang and Liu, Liyuan and Li, Sha and Han, Jiawei},
  booktitle={Proceedings of the web conference 2020},
  pages={1908--1919},
  year={2020}
}

@article{hsu2024chime,
  title={Chime: Llm-assisted hierarchical organization of scientific studies for literature review support},
  author={Hsu, Chao-Chun and Bransom, Erin and Sparks, Jenna and Kuehl, Bailey and Tan, Chenhao and Wadden, David and Wang, Lucy Lu and Naik, Aakanksha},
  journal={Findings of the Association for Computational Linguistics: ACL 2024},
  pages={118--132},
  year={2024}
}

@inproceedings{kargupta2025taxoadapt,
  title={Taxoadapt: Aligning llm-based multidimensional taxonomy construction to evolving research corpora},
  author={Kargupta, Priyanka and Zhang, Nan and Zhang, Yunyi and Zhang, Rui and Mitra, Prasenjit and Han, Jiawei},
  booktitle={Proceedings of the 63rd Annual Meeting of the Association for Computational Linguistics (Volume 1: Long Papers)},
  pages={29834--29850},
  year={2025}
}

@article{mcinnes2017hdbscan,
  title={hdbscan: Hierarchical density based clustering.},
  author={McInnes, Leland and Healy, John and Astels, Steve and others},
  journal={J. Open Source Softw.},
  volume={2},
  number={11},
  pages={205},
  year={2017}
}

@inproceedings{katz2024knowledge,
  title={Knowledge navigator: Llm-guided browsing framework for exploratory search in scientific literature},
  author={Katz, Uri and Levy, Mosh and Goldberg, Yoav},
  booktitle={Findings of the Association for Computational Linguistics: EMNLP 2024},
  pages={8838--8855},
  year={2024}
}

@inproceedings{zeng2024chain,
  title={Chain-of-layer: Iteratively prompting large language models for taxonomy induction from limited examples},
  author={Zeng, Qingkai and Bai, Yuyang and Tan, Zhaoxuan and Feng, Shangbin and Liang, Zhenwen and Zhang, Zhihan and Jiang, Meng},
  booktitle={Proceedings of the 33rd ACM International Conference on Information and Knowledge Management},
  pages={3093--3102},
  year={2024}
}

@article{wullschleger2025no,
  title={No gold standard, no problem: Reference-free evaluation of taxonomies},
  author={Wullschleger, Pascal and Zarharan, Majid and Daly, Donnacha and Pouly, Marc and Foster, Jennifer},
  journal={arXiv preprint arXiv:2505.11470},
  year={2025}
}

@article{li2026grading,
  title={Grading Scale Impact on LLM-as-a-Judge: Human-LLM Alignment Is Highest on 0-5 Grading Scale},
  author={Li, Weiyue and Zhao, Minda and Dong, Weixuan and Cai, Jiahui and Wei, Yuze and Pocress, Michael and Li, Yi and Yuan, Wanyan and Wang, Xiaoyue and Hou, Ruoyu and others},
  journal={arXiv preprint arXiv:2601.03444},
  year={2026}
}

@article{hurst2024gpt,
  title={Gpt-4o system card},
  author={Hurst, Aaron and Lerer, Adam and Goucher, Adam P and Perelman, Adam and Ramesh, Aditya and Clark, Aidan and Ostrow, AJ and Welihinda, Akila and Hayes, Alan and Radford, Alec and others},
  journal={arXiv preprint arXiv:2410.21276},
  year={2024}
}

\appendix
\section*{Appendix}

\section{Algorithm}
\label{app:pseudo_code}
Algorithm~\ref{alg:evotaxo} summarizes the full EvoTaxo pipeline.
\begin{algorithm}[t]
\caption{EvoTaxo}
\label{alg:evotaxo}
\small
\begin{algorithmic}[1]
\Require temporally ordered posts $\mathcal{D}=\{(x_i,\tau_i)\}_{i=1}^N$, root node $r$, time windows $\mathcal{W}=\{W_1,\dots,W_M\}$
\Ensure taxonomy $\mathcal{T}$ and temporal grounding records $\Gamma$
\State $\mathcal{T} \gets \textsc{SeedTaxonomy}(r)$
\State $\mathcal{B} \gets \emptyset$, $\Gamma \gets \emptyset$
\For{each time window $W_k \in \mathcal{W}$}
    \State $\mathcal{C}_k \gets \emptyset$, $\mathcal{R}_k \gets \emptyset$
    \For{each post $(x_i,\tau_i)$ such that $\tau_i \in W_k$}
        \State $a_i \gets f_{\mathrm{prop}}(x_i, \mathcal{T})$
        \If{$a_i$ is structural}
            \State $\mathcal{B} \gets \mathcal{B} \cup \{a_i\}$
        \Else
            \State $(\mathcal{T}, \Gamma) \gets \textsc{ExecuteImmediate}(a_i, \mathcal{T}, \Gamma)$
        \EndIf
    \EndFor
    \State $\{\mathcal{G}_1,\dots,\mathcal{G}_J\} \gets \textsc{PartitionBacklog}(\mathcal{B})$
    \For{each bucket $\mathcal{G}_j$}
        \State $\mathcal{C}_k \gets \mathcal{C}_k \cup \textsc{DualViewCluster}(\mathcal{G}_j)$
    \EndFor
    \For{each candidate cluster $C \in \mathcal{C}_k$}
        \State $\mathcal{R}_k \gets \mathcal{R}_k \cup \textsc{RefineCluster}(C, \mathcal{T})$
    \EndFor
    \State $\mathcal{F}_k \gets \textsc{Arbitrate}(\mathcal{R}_k, \mathcal{T})$
    \State $(\mathcal{T}, \Gamma, \mathcal{B}) \gets \textsc{ApplyFinalActions}(\mathcal{F}_k, \mathcal{T}, \Gamma, \mathcal{B})$
\EndFor
\State \Return $\mathcal{T}, \Gamma$
\end{algorithmic}
\end{algorithm}

\section{Data Preprocessing}
\label{app:data}

We collect Reddit data from \texttt{/r/opiates} and \texttt{/r/ICE\_Raids} using Arctic Shift.\footnote{\url{https://arctic-shift.photon-reddit.com/}} For each subreddit, we retrieve posts together with associated discussion content and construct the text units used in our experiments. We then use an NLI model (\texttt{facebook/bart-large-mnli}) as a coarse filtering tool to separate broad discourse types, such as viewpoint exchange, advice seeking, and information sharing. Our purpose is to retain content that is more likely to contain substantive opinion-oriented discussion while filtering out lower-value or less relevant material for taxonomy construction. After scoring and sampling, we keep viewpoint-exchange data whose confidence score exceeds $0.75$. Data preprocessing is applied to all baseline methods.

\section{Experimental Settings}
\label{app:exp_details}

Unless otherwise noted, we use the same hyperparameter settings across all datasets. All LLM calls use temperature $0.0$. For EvoTaxo's time-aware clustering view, we assign equal importance to semantic distance and normalized temporal distance, with $\lambda = 0.5$. For both the semantic and temporal HDBSCAN passes, we use \texttt{min\_cluster\_size}=10, meaning that at least ten proposals are required before a group can be treated as a non-noise cluster candidate. We do not manually tune \texttt{min\_samples}; instead, we leave it at the implementation default. 

We compare EvoTaxo against four representative baselines. \textbf{Knowledge Navigator} (KN)~\cite{katz2024knowledge} combines clustering with LLM-based hierarchy construction: it first groups documents using clustering-style organization and then asks the LLM to interpret and name the resulting structure. \textbf{Chain-of-Layer}~\cite{zeng2024chain} constructs a taxonomy from a provided entity set through iterative layer-wise prompting and filtering; in other words, it starts from candidate entities rather than directly inducing structure from post-level corpus interaction. We follow the implementation described in TaxoAdapt~\cite{kargupta2025taxoadapt} to provide the entity set. \textbf{TnT-LLM}~\cite{wan2024tnt} is implemented following the Delve framework: its taxonomy-construction stage first summarizes sampled texts and then iteratively refines the taxonomy over those summaries, after which taxonomy assignment is performed as a separate stage. \textbf{TaxoAdapt}~\cite{kargupta2025taxoadapt} starts from an LLM-generated taxonomy and repeatedly adapts it to the corpus through iterative hierarchical classification and taxonomy expansion, requiring multiple rounds of corpus-level optimization.

\section{Evaluation Prompts}
\label{app:eval_prompts}

Below we show the prompt templates used for the three LLM-based structural evaluation metrics in Section~\ref{sec:exp_setup}. In each case, placeholders such as \texttt{<ROOT\_TOPIC>}, \texttt{<PARENT>}, \texttt{<PATH>}, and \texttt{<SIBLINGS>} are replaced with the corresponding taxonomy content.

\paragraph{Path Granularity:}
\begin{quote}
\small
\ttfamily
You are evaluating path granularity in a \texttt{<ROOT\_TOPIC>} taxonomy.

The path is: \texttt{<PATH>}.

Score whether this path becomes progressively more specific from parent to child. A high score means each child is a clear and sensible refinement of its parent. A low score means the path has jumps in abstraction, inconsistent granularity, or overly broad/narrow transitions. Use a 1--5 scale where 1 = very poor granularity, 2 = poor, 3 = mixed/acceptable, 4 = good, and 5 = excellent granularity. Return a short rationale and an explicit machine-readable score in the form '<score: X>'.
\end{quote}

\paragraph{Sibling Coherence:}
\begin{quote}
\small
\ttfamily
\noindent You are evaluating sibling coherence in a \texttt{<ROOT\_TOPIC>} taxonomy.

\noindent The parent topic is: \texttt{<PARENT>}.

\noindent The sibling categories are: \texttt{<SIBLINGS>}.

\noindent Score how coherent this sibling set is as a group. A high score means the siblings belong together conceptually and operate at a comparable level of abstraction relative to the parent. A low score means the set feels mixed, uneven, or poorly grouped. Use a 1--5 scale where 1 = very incoherent, 2 = weak, 3 = mixed/acceptable, 4 = good, and 5 = very strong coherence. Return a short rationale and an explicit machine-readable score in the form '<score: X>'.
\end{quote}

\paragraph{Sibling Separability:}
\begin{quote}
\small
\ttfamily
\noindent You are evaluating sibling separability in a \texttt{<ROOT\_TOPIC>} taxonomy.

\noindent The parent topic is: \texttt{<PARENT>}.

\noindent The sibling categories are: \texttt{<SIBLINGS>}.

\noindent Score how clearly these sibling categories are distinguishable from one another. A high score means each sibling has a clear, non-overlapping scope and could be reliably told apart from the others. A low score means the siblings are confusing, redundant, or strongly overlapping. Use a 1--5 scale where 1 = very poor separability, 2 = weak, 3 = mixed/acceptable, 4 = good, and 5 = excellent separability. Return a short rationale and an explicit machine-readable score in the form '<score: X>'.
\end{quote}

\section{LLM-Human Agreement Analysis}
\label{app:agreement}

Since part of our evaluation suite relies on LLM-based structural judgment, we conduct a small-scale human evaluation to assess the reliability of these metrics. We focus on the three LLM-judged metrics used in Section~\ref{exp}: \textit{Path Granularity}, \textit{Sibling Coherence}, and \textit{Sibling Separability}.

We randomly sample evaluation instances from the taxonomies produced by all compared methods across datasets. For \textit{Path Granularity}, we sample 30 root-to-leaf paths. For \textit{Sibling Coherence} and \textit{Sibling Separability}, we each sample 30 sibling sets. A human annotator independently evaluates each instance using the same rubric as the corresponding LLM prompt in Appendix~\ref{app:eval_prompts}, assigning a score on the same 1--5 scale.

We then compare the human scores with those produced by the LLM judge (\texttt{GPT-4o-mini}). Because these metrics are ordinal, we report both exact agreement and relaxed agreement within $\pm 1$ point. The results are shown in Table~\ref{tab:agreement}. Overall, the agreement levels indicate that the LLM judge is reasonably well aligned with human judgment, supporting the use of these metrics as scalable proxies for taxonomy structural quality.

\begin{table}[h]
\centering
\small
\caption{LLM-human agreement for the three LLM-judged structural metrics.}
\label{tab:agreement}
\setlength{\tabcolsep}{9pt}
\begin{tabular}{lccc}
\toprule
\textbf{Metric} & \textbf{Sample Size} & \textbf{Exact Agr.} & \textbf{$\pm 1$ Agr.} \\
\midrule
Path & 30 & 0.67 & 0.90 \\
Sib-C & 30 & 0.70 & 0.87 \\
Sib-S & 30 & 0.57 & 0.83 \\
\bottomrule
\end{tabular}
\end{table}

\section{Appendix Case Study}

\begin{figure*}[t!]
    \centering
    \includegraphics[width=\textwidth]{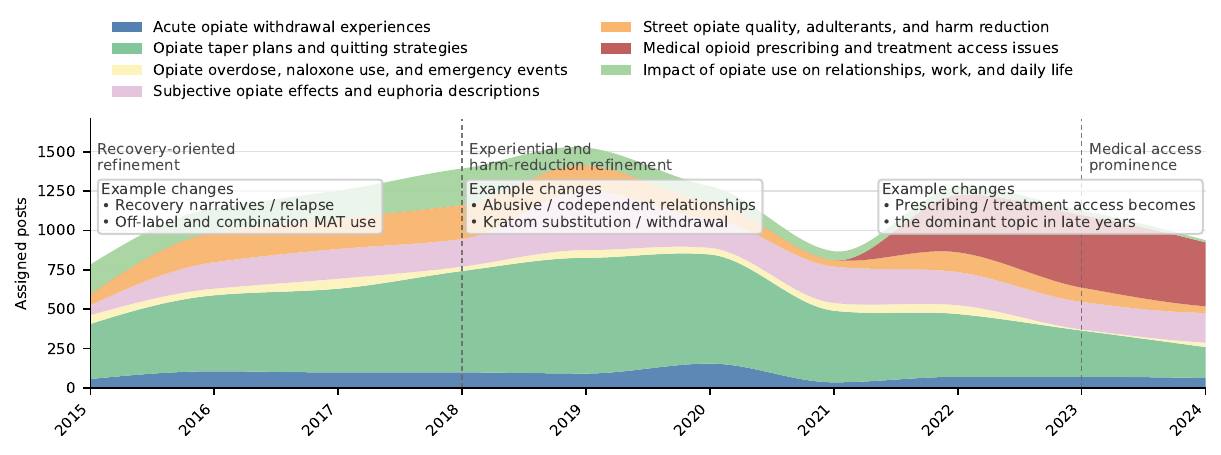}
    \vspace{-8pt}
    \caption{Yearly evolution of the taxonomy induced from \texttt{/r/opiates}, aggregated to the seven top-level topics. Early years are dominated by tapering/quitting and street-opiate quality and harm-reduction discussion. Later years show a stronger shift toward medical prescribing and treatment access.}
    \label{fig:opiates_case_study}
    \vspace{-8pt}
\end{figure*}

\subsection{Temporal Micro-Bursts on \texttt{/r/ICE\_Raids}}
\label{app:ice_microbursts}

The main-text ICE case study highlights large month-level transitions, but the temporal view is also valuable at a finer scale. In several instances, EvoTaxo introduces subtopics whose total support is modest in absolute terms, yet whose evidence is concentrated within a few days. These cases are precisely the patterns most likely to be diluted under corpus-wide semantic grouping, since they do not require large aggregate volume to constitute a coherent structural development.

One example is \emph{ICE coordination with private individuals and entities for enforcement}. This subtopic accumulates only around 50 supporting posts overall, but 36 of them occur on a single day in November 2025. Another example is \emph{ICE deportation record-keeping and transparency issues}, which receives 53 of its 67 supporting posts within one week in May 2025. A third case is \emph{ICE enforcement at immigration courts and civil proceedings}, for which 64 of 112 supporting posts arrive within a one-week span centered at the end of May 2025. Although these subtopics are not dominant at the monthly aggregate level, each reflects a temporally compact and semantically coherent discussion episode.

These examples sharpen the interpretation of EvoTaxo's temporal component. The value of time-aware clustering is not limited to capturing large protest waves or major policy surges; it also helps surface low-volume, short-lived developments that nonetheless warrant explicit representation in the taxonomy. In this sense, the temporal view acts as a sensitivity mechanism for emerging discourse fragments that are locally dense in time even when they are globally sparse in the corpus.

\subsection{Yearly Evolution on \texttt{/r/opiates}}
\label{app:opiates_case}

In contrast to the event-driven \texttt{/r/ICE\_Raids} case in the main paper, the yearly evolution of the \texttt{/r/opiates} taxonomy is characterized by gradual refinement over a relatively stable high-level structure. The top-level organization is established early and remains largely intact across the full 2015--2024 horizon. Rather than repeatedly reorganizing the taxonomy around sudden bursts, EvoTaxo mainly refines persistent community concerns as they accumulate sufficient support over multiple years, as illustrated in Figure~\ref{fig:opiates_case_study}.

The earliest years are dominated by two broad themes. First, \emph{Opiate taper plans and quitting strategies} becomes the most consistently active topic across much of the timeline, indicating that quitting, tapering, and self-managed recovery are central organizing concerns. Second, \emph{Street opiate quality, adulterants, and harm reduction} is especially prominent in the early period, reflecting sustained attention to supply quality, contamination, potency, and risk reduction. This combination suggests that the community is initially structured around a practical mixture of survival knowledge, self-management, and street-use realities.

The structural updates introduced by EvoTaxo are correspondingly selective. In 2015, the taxonomy adds subtopics related to recovery narratives and off-label or combination use of medication-assisted treatment, sharpening the distinction between general tapering talk and more specific treatment-oriented quitting strategies. In 2016 and 2018, the taxonomy further refines recurring experiential themes by separating relationship burden, overdose-related family impact, street-safety incidents, and later kratom-related substitution or withdrawal discussion. These additions do not reflect abrupt discourse shocks; instead, they indicate that recurring motifs become stable enough over time to justify explicit branches in the hierarchy.

Another notable pattern is that later refinements become more specialized. By 2019, the taxonomy introduces a subtopic for drug-specific subjective experiences and first-time expectations, indicating that the discussion is no longer captured only by broad euphoria or withdrawal categories. Rather, the community develops a more differentiated vocabulary for comparing effects, substances, and use trajectories. This kind of delayed refinement is consistent with a long-running discussion space in which distinctions emerge through repeated experiential reporting rather than through short-lived external events.

At the topic-distribution level, the later years also show a gradual shift toward medical access and treatment systems. \emph{Medical opioid prescribing and treatment access issues} becomes substantially more prominent in 2022--2024. This suggests that the center of discussion increasingly moves toward prescription access, clinical interaction, and treatment navigation. Taken together, the yearly \texttt{/r/opiates} case study highlights a different mode of taxonomy evolution from the ICE example: instead of temporally concentrated bursts, EvoTaxo captures slow semantic differentiation within a stable community discourse structure.

\end{document}